\documentclass[sigconf]{acmart}
\AtBeginDocument{%
  }

\acmISBN{978-1-4503-XXXX-X/2018/06}

\usepackage{subcaption}
\usepackage{booktabs}   
\usepackage{multirow}   
\usepackage{tcolorbox}
\usepackage{xcolor}
\usepackage{enumitem}
\usepackage{amsmath}
\usepackage{tabularx}
\usepackage{array}
\usepackage{tikz}
\usepackage{adjustbox}

\usetikzlibrary{positioning,fit,calc,arrows.meta,shapes.misc}
\usepackage{caption}
\usepackage{tcolorbox}          
\tcbuselibrary{breakable}       

\newtcolorbox{casebox}[2][]{breakable,
  colback=white, colframe=black, boxrule=0.35pt, arc=2pt,
  left=6pt, right=6pt, top=6pt, bottom=6pt,
  fonttitle=\bfseries, title={#2}, #1}

\definecolor{darkblue}{rgb}{0.0, 0.0, 0.55}
\definecolor{forestgreen}{rgb}{0.0, 0.27, 0.13}
\definecolor{tomato}{rgb}{1.0, 0.39, 0.28}
\definecolor{burgundy}{rgb}{0.5, 0.0, 0.13}
\definecolor{apple}{rgb}{0.55, 0.71, 0.0}
\definecolor{mediumviolet}{rgb}{0.78, 0.08, 0.52} 
\definecolor{red}{rgb}{0.77, 0.01, 0.2}

\usepackage{booktabs,xcolor,colortbl,pgf} 

\definecolor{AccGreen}{RGB}{49,163,84}   
\definecolor{GainBlue}{RGB}{33,113,181}  
\definecolor{PenRed}{RGB}{204,24,30}     

\newcommand{\AccCell}[1]{%
  \pgfmathsetmacro{\accpct}{min(100,max(0,100*(#1)))}%
  \edef\colspec{AccGreen!\accpct!white}%
  \expandafter\cellcolor\expandafter{\colspec}#1%
}
\newcommand{\DelCell}[1]{
  \pgfmathsetmacro{\delpct}{min(100,max(0,300*(#1)))}%
  \edef\colspec{GainBlue!\delpct!white}%
  \expandafter\cellcolor\expandafter{\colspec}#1%
}
\newcommand{\PenCell}[1]{
  \pgfmathsetmacro{\penmag}{min(100,max(0,400*abs(#1)))}%
  \ifdim #1pt<0pt
    \edef\colspec{PenRed!\penmag!white}%
  \else
    \edef\colspec{AccGreen!\penmag!white}%
  \fi
  \expandafter\cellcolor\expandafter{\colspec}#1%
}

\begin{document}

\title{i\textsc{AgentBench}: Benchmarking Sensemaking Capabilities of Information-Seeking Agents on High-Traffic Topics}

\author{Preetam Prabhu Srikar Dammu}
\affiliation{
    \institution{University of Washington}
    \city{Seattle}
    \state{Washington}
    \country{USA}
}
\email{preetams@uw.edu}

\author{Arnav Palkhiwala}
\affiliation{
    \institution{University of Washington}
    \city{Seattle}
    \state{Washington}
    \country{USA}
}
\email{apalkh@uw.edu}

\author{Tanya Roosta}
\affiliation{
    \institution{UC Berkeley}
    \city{Berkeley}
    \state{California}
    \country{USA}
}
\email{troosta@ischool.berkeley.edu}

\author{Chirag Shah}
\affiliation{
    \institution{University of Washington}
    \city{Seattle}
    \state{Washington}
    \country{USA}
}
\email{chirags@uw.edu}

\renewcommand{\shortauthors}{Dammu et al.}

\begin{abstract}
With the emergence of search-enabled generative QA systems, users are increasingly turning to tools that browse, aggregate, and reconcile evidence across multiple sources on their behalf. Yet many widely used QA benchmarks remain answerable by retrieving a single relevant passage, making them poorly suited for measuring cross-source sensemaking, such as integrating evidence, tracking causal links, and resolving dependencies across facets of a topic. We present i\textsc{AgentBench}, a dynamic ODQA benchmark that targets these higher-level information needs while keeping questions natural and grounded in realistic information-seeking behavior. i\textsc{AgentBench} draws seed topics from real-world attention signals and uses common user intent patterns to construct user-like questions whose answers require combining evidence from multiple sources, not just extracting a single snippet. Each instance is released with traceable evidence and auditable intermediate artifacts that support contamination checks and enable fine-grained diagnosis of failures in retrieval versus synthesis. 
Experiments across multiple LLMs show that retrieval improves accuracy, but retrieval alone does not reliably resolve these questions, underscoring the need to evaluate evidence use, not just evidence access.
\end{abstract}





\maketitle

\section{Introduction}

Information-seeking agents (ISAs) are increasingly used when a user needs more than a quick lookup. In these settings, the system must search across sources, reconcile partial evidence, and produce a concise synthesis that supports a downstream decision. When an answer is contained in a single passage, strong retrieval plus generation is often enough. When the answer only emerges after integrating multiple sources, reliability hinges on how well the system accumulates, organizes, and uses evidence.

Yet evaluation still over-indexes on questions that collapse to finding one relevant passage. Many widely used QA benchmarks were designed around passage-centric answering, where success is dominated by matching and extracting the right span or sentence, even under adversarial perturbations \cite{jiaAdversarialExamplesEvaluating2017}. This setup is a weak proxy for the cross-source understanding that motivates ISAs. It under-measures failure modes that arise when evidence is distributed across documents, when multiple themes must be considered jointly, or when the correct response depends on how those themes relate.

A natural response is to turn to more challenging QA settings, particularly multi-hop QA. However, multi-hop reasoning should not be conflated with sensemaking, which requires capabilities beyond chaining evidence across hops. Datasets such as HotPotQA still often reward locating a small number of lexical matches or short supporting statements \cite{yangHotpotQADatasetDiverse2018a}. Even when multiple documents are involved, many multi-hop questions are structured so that the work is primarily path-following and snippet stitching. 

The notion of \emph{sensemaking questions}, as defined by \citet{edge2024local}, helps clarify what is missing. \citet{edge2024local} argue that conventional RAG struggles on queries that require query-focused summarization of an entire text collection, rather than retrieving a few matching records. This dataset-level framing is valuable, but it targets a different regime than most open-domain ISAs. In practice, the relevant context is rarely a fixed private corpus. It is a query-conditioned slice of a large and changing web, where reading the first few pages of search results is often the natural scope for answering the user. Evaluating agents in this regime calls for benchmarks that focus on sensemaking over retrieved sources themselves, not global collection-level summarization.

We introduce i\textsc{AgentBench}, short for \emph{Information-seeking Agent Benchmark}, a dynamic benchmark that targets \emph{retrieved-sources-level sensemaking} for open-domain QA. Each instance is seeded from traffic-driven, time-indexed topics so it reflects what users were actively seeking, and grounded in a query-conditioned slice of open-web evidence. Questions are instantiated via intent patterns to stay natural and standalone, while the story-graph representation makes themes and connector relations explicit so that answering requires integrating evidence across multiple themes, not extracting a single passage or performing superficial multi-hop stitching.

i\textsc{AgentBench} is designed to support repeated evaluation as the web changes and to enable analysis beyond headline accuracy by distinguishing failures of evidence access from failures of evidence integration.
Specifically, i\textsc{AgentBench} is designed with the following benchmark-level characteristics:
\begin{enumerate}\setlength{\itemsep}{0pt}
    \item \emph{Traffic-driven, time-indexed topics} grounded in real-world attention signals.
    \item \emph{Query-conditioned open-web corpora} that reflect the bounded evidence slice an ISA would read at test time.
    \item \emph{Cross-theme, connector-dependent questions} with enforced dependence on multiple themes and their relations (not single-passage lookup or superficial multi-hop stitching).
    \item \emph{Questions that reflect user informational needs} by instantiating common intent patterns, so evaluation tests sensemaking goals beyond generic fact lookup.
    \item \emph{Dynamic regeneration over time windows} to support repeated evaluation on fresh evidence and reduce exposure to memorization.
\end{enumerate}

We make i\textsc{AgentBench} available on Hugging Face\footnote{\url{https://huggingface.co/datasets/preetam7/iAgentBench}}, the source code on GitHub\footnote{\url{https://github.com/iagentbench/iAB.git}}, and additional resources on the project website\footnote{\url{https://iagentbench.github.io/iAgentBench/}} to support research on evaluating information-seeking agents under dynamic, open-web conditions.

\section{Related Work}
\label{sec:related_work}

Open-domain question answering (ODQA) benchmarks differ substantially in how closely their questions reflect real user intent.
Reading-comprehension datasets such as SQuAD \cite{rajpurkarSQuAD100000Questions2016} helped standardize span extraction, while later ODQA resources moved toward questions that resemble genuine web information needs.
Natural Questions \cite{kwiatkowskiNaturalQuestionsBenchmark2019a} is grounded in real search queries, and (QA)$^2$ \cite{kimQA$^2$QuestionAnswering2023} highlights that naturally occurring questions can include questionable assumptions.
At the same time, many widely used ODQA datasets are shaped by benchmark convenience or authoring constraints (e.g., trivia-oriented sources), which can skew the distribution away from questions users actually ask.
In this work, i\textsc{AgentBench} follows an interest-driven design: seed topics are drawn from real-world attention signals (via GDELT~\cite{gdeltproject}) rather than curated knowledge bases or quiz-style collections.

A central challenge in ODQA is that evidence is often distributed across multiple documents, and relevant evidence may not be semantically similar to the question itself.
HotpotQA \cite{yangHotpotQADatasetDiverse2018a} popularized multi-hop supervision, and retrieval-heavy readers such as FiD \cite{izacardLeveragingPassageRetrieval2021} show the gains from aggregating many passages.
Retriever development has similarly focused on improving coverage and compositionality, including dense retrieval \cite{karpukhinDensePassageRetrieval2020b} and retrieval-augmented language models \cite{lewisRetrievalAugmentedGenerationKnowledgeIntensive2020a,guuRetrievalAugmentedLanguage2020,izacardAtlasFewshotLearning2022}.
These benchmarks and systems are essential for studying retrieval quality and multi-hop composition.
By contrast, i\textsc{AgentBench} shapes questions to require \emph{cross-theme synthesis} over a query-conditioned web corpus, where the bottleneck is often not a single missing passage but integrating evidence that is dispersed across themes and their explicit linkages.

Graph structure has been used to support multi-step retrieval and to make cross-document reasoning more explicit.
Early work retrieves reasoning paths over Wikipedia graphs for multi-hop QA \cite{asaiLearningRetrieveReasoning2020a}, reflecting the value of structured traversal when lexical overlap is limited.
More recently, GraphRAG-style \cite{edge2024local} approaches propose building intermediate graph representations from a corpus to enable higher-level synthesis, often relying on community discovery methods such as Leiden~\cite{traagLouvainLeidenGuaranteeing2019}.
In particular, ``local-to-global'' query-focused summarization highlights that certain questions benefit from global structure over the relevant corpus rather than purely local passage matching \cite{edge2024local}.
i\textsc{AgentBench} builds on these insights but targets a different setting: instead of dataset-level sensemaking over a fixed corpus, we construct \emph{query-conditioned} story graphs over web-retrieved evidence and generate ODQA pairs that implicitly require integrating multiple themes and their explicit inter-theme connectors.

\begin{figure*}[t]
    \centering
    \includegraphics[width=0.75\textwidth]{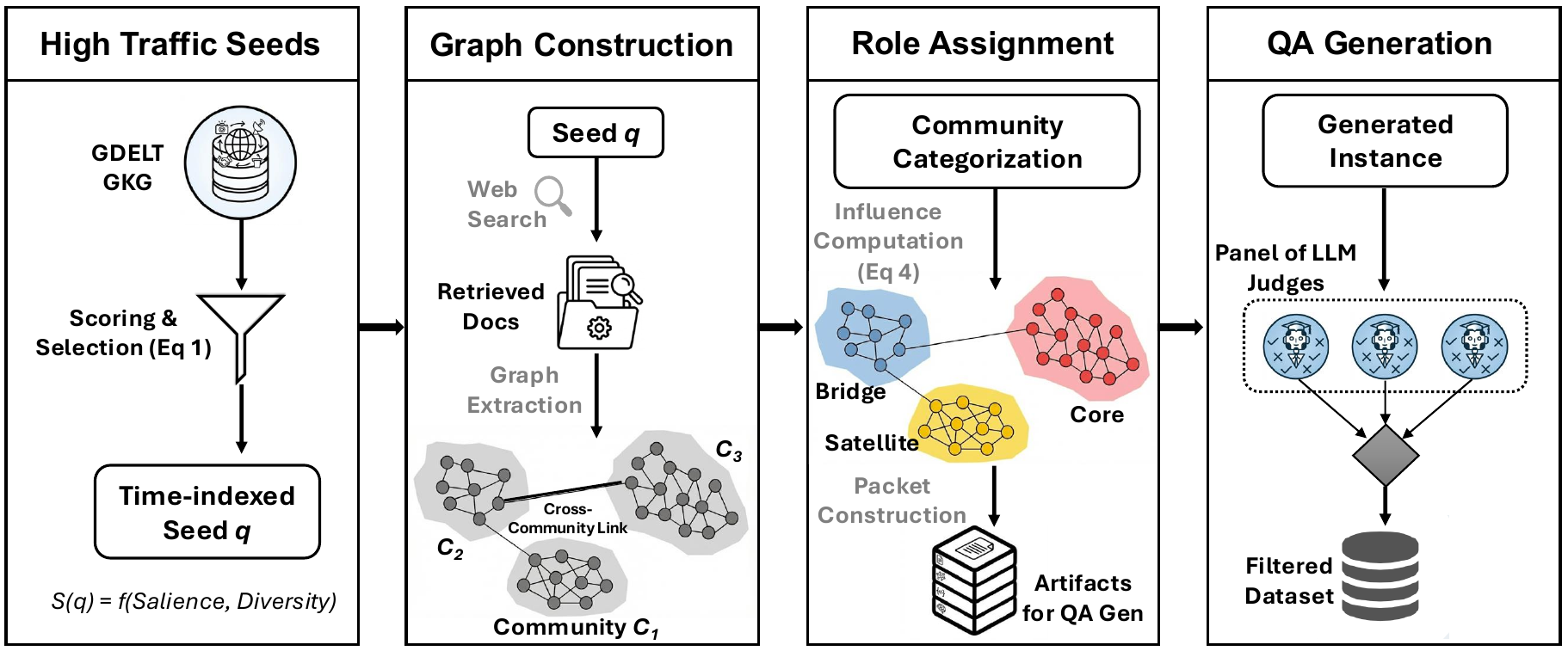}
    \caption{Overview of the i\textsc{AgentBench} construction pipeline. We (1) sample time-indexed, high-traffic seed queries from public attention signals (GDELT), (2) retrieve a query-conditioned web corpus and extract a claim-like story graph with thematic communities, (3) assign community roles (\textsc{Core}/\textsc{Bridge}/\textsc{Satellite}) and build compact artifacts (community cards, connectors, packets) that preserve cross-theme links, and (4) generate and filter standalone ODQA pairs using a panel of LLM judges.}
    \label{fig:pipeline}
\end{figure*}

As QA systems increasingly incorporate tools (search, browsing, and iterative planning), evaluation must stress behaviors beyond one-shot retrieval.
WebGPT \cite{nakanoWebGPTBrowserassistedQuestionanswering2022a} and ReAct \cite{yaoReActSynergizingReasoning2023a} exemplify systems that interleave evidence gathering with generation.
This trend also amplifies longstanding benchmarking concerns: static QA datasets age quickly and are increasingly susceptible to memorization and leakage as model training corpora expand \cite{carliniQuantifyingMemorizationNeural2023,liOpenSourceData2024,ravautComprehensiveSurveyContamination2025a}.
Recent work has therefore emphasized time-sensitive evaluations \cite{kasaiRealTimeQAWhats2024,liskaStreamingQABenchmarkAdaptation2022} and dynamic benchmarking frameworks \cite{kielaDynabenchRethinkingBenchmarking2021a,maDynaboardEvaluationAsAServicePlatform2021,rawlesAndroidWorldDynamicBenchmarking2025,zhuDyVal2Dynamic2024,zhangDARGDynamicEvaluation2024,dammuDynamicKGQAScalableFramework2025a}.
In retrieval-augmented settings, search-time contamination is an additional failure mode, where agents can directly retrieve benchmark content and copy answers \cite{hanSearchTimeDataContamination2025}.
Accordingly, i\textsc{AgentBench} is designed to be refreshable and auditable: it is anchored in time-indexed, interest-driven topic signals and produces evidence artifacts (retrieved documents, story-graph links, and annotations) that support contamination checks and qualitative error analysis.
\section{Method}
\label{sec:methods}

i\textsc{AgentBench} is a dynamic, open-domain QA benchmark construction pipeline designed to evaluate \emph{cross-document sensemaking} for information-seeking agents. The key idea is to start from realistic, traffic-driven topics, retrieve a query-conditioned corpus from the web, build a compact structured representation of how themes in that corpus relate, and then generate questions whose answers depend on multiple themes and the explicit links between them. Figure~\ref{fig:pipeline} summarizes the pipeline.

\subsection{Interest-Driven Seeds}
\label{sec:seed-queries}

A central goal of i\textsc{AgentBench} is to support QA that users plausibly care about, rather than questions induced by artifacts of a fixed source database (e.g., a knowledge base schema or a curated trivia list). We therefore derive seed topics from public attention signals and time-index them to enable dynamic evaluation.

\paragraph{Seed candidates from GDELT.}
We use daily snapshots from the GDELT 1.0 Global Knowledge Graph (GKG) \cite{gdeltproject} as a proxy for what events and entities are receiving attention. Each day yields a set of candidate event-style seeds by combining salient entities with event descriptors (e.g., entity + action phrase) extracted from GKG records. Aggregating over a time window produces a pool of candidates that naturally reflects changing interests.

 \paragraph{Scoring and selection.}
 Let $\mathcal{Q}$ denote all candidate seeds aggregated over a window. For each $q\in\mathcal{Q}$ we compute a score that balances salience, temporal specificity, and diversity:
\begin{align}
\label{eq:seed-score}
S(q)\;=&\;\log(1+A(q))
\; +\; \alpha\,\mathrm{Geo}(q)
\; +\; \beta\,\mathrm{Freq}(q)\nonumber\\
&\; +\; \gamma\,\Delta t(q)
\; +\; \delta\,\mathrm{Spec}(q)
\; -\; \eta\,\mathrm{Cov}(q).
\end{align}
Here $A(q)$ is an attention proxy when available (otherwise a monotone function of daily observations), $\mathrm{Geo}(q)$ is geographic breadth, $\mathrm{Freq}(q)$ is observation count, $\Delta t(q)$ is the date span covered by the signal, $\mathrm{Spec}(q)$ is a light specificity bonus (to avoid generic seeds), and $\mathrm{Cov}(q)$ penalizes candidates that occur nearly every day (often indicating generic terms).

\subsection{Graph Construction}
\label{sec:story-graph}

Given a selected seed query $q$, we retrieve a query-conditioned corpus $\mathcal{D}(q)=\{d_i\}_{i=1}^{N}$ from web sources using a standard search API. This is the evaluation setting we target: an agent that can search, read, and synthesize from a small set of relevant sources returned at test time. In contrast to dataset-level summarization settings, the benchmark does not assume (or attempt) a persistent global representation of an entire corpus such as ``the web.''

To support cross-document sensemaking, we then construct a query-specific structured representation that aggregates information across $\mathcal{D}(q)$. Following GraphRAG-style pipelines \cite{edge2024local}, we use an LLM-assisted extractor to identify entities and relational assertions from text. The resulting structure is not a schema-fixed knowledge graph. Relations are short natural-language claims grounded in evidence, which makes the representation closer to a hypergraph of assertions than a predicate-limited KG.

\paragraph{Graph definition.}
We construct a graph
\begin{equation}
\label{eq:story-graph}
G(q) = (V, E),
\end{equation}
where each node $v\in V$ corresponds to an entity, and each edge $e\in E$ is a tuple $(s, r, o, \mathrm{ev})$ with subject $s\in V$, object $o\in V$, relation text $r$ (a claim-like statement), and evidence pointers $\mathrm{ev}\subseteq\mathcal{U}$ to underlying text units (sentences or chunks). Each edge also carries a support signal (e.g., edge weight, frequency, or evidence mass).

\paragraph{Community detection and reports.}
We partition $G(q)$ into communities $\mathcal{C}(q)=\{c_1,\ldots,c_m\}$ using Leiden clustering. We interpret each community as a \emph{theme} (a coherent sub-story within the retrieved corpus). For each community $c$, the pipeline produces a short natural-language summary and a set of grounded findings (salient statements tied to evidence). These artifacts provide a compact, human-readable interface to the thematic structure of $\mathcal{D}(q)$.

\subsection{Community Roles and Influence}
\label{sec:community-roles}

 A major challenge in turning $G(q)$ into a benchmark is scale and noise. Large graphs contain many peripheral entities and redundant edges, and not all themes are equally important for question generation. To focus question construction on the most consequential parts of the corpus, we compute a theme-level view of the graph and assign each community a \emph{role} that guides downstream selection.

\paragraph{Community meta-graph.}
We define a community meta-graph $H(q)=(\mathcal{C}(q), E_H)$ in which each node is a community and edges aggregate cross-community relations induced by $G(q)$.
\begin{equation}
\label{eq:meta-graph}
H(q)=(\mathcal{C}(q), E_H).
\end{equation}
Intuitively, $H(q)$ captures how themes interact within the query-conditioned corpus. The weight of an edge $(c_i,c_j)\in E_H$ reflects both the strength and the quantity of cross-community evidence.

 \paragraph{Influence score.}
 We compute an influence score $I(c)$ for each community that combines size, connectivity, and evidence support:
\begin{align}
\label{eq:influence}
I(c)\;=&\;\tfrac{1}{4}\,z\!\left(\log(1+|c|)\right)
\; +\; \tfrac{1}{4}\,z\!\left(\operatorname{PR}(c)\right)\nonumber\\
&\; +\; \tfrac{1}{4}\,z\!\left(\operatorname{BC}(c)\right)
\; +\; \tfrac{1}{4}\,z\!\left(\operatorname{Ev}(c)\right),
\end{align}
where $|c|$ is community size, $\operatorname{PR}$ and $\operatorname{BC}$ are PageRank and betweenness centrality on $H(q)$, $\operatorname{Ev}(c)$ counts unique evidence units supporting $c$, and $z(\cdot)$ denotes within-query standardization.

\paragraph{Core, Bridge, and Satellite themes.}
We assign each community a role using lightweight, query-local rules derived from the meta-graph $H(q)$.
Intuitively, \textsc{Core} captures dominant themes, \textsc{Bridge} captures themes that mediate between otherwise separate themes, and \textsc{Satellite} captures peripheral themes that attach strongly to core/bridge structure.
\begin{itemize}
    \item \textsc{Core}: high-influence themes that represent dominant sub-stories in $\mathcal{D}(q)$.
    \item \textsc{Bridge}: themes with high betweenness that connect otherwise separate sub-stories, often mediating how evidence in one theme affects another.
    \item \textsc{Satellite}: peripheral themes that provide supporting context and attach strongly to a core or bridge.
\end{itemize}
Concretely, let $I(c)$ be the influence score in Eq.~\eqref{eq:influence} and let $\operatorname{BC}(c)$ denote betweenness centrality on $H(q)$. We define a simple selection operator $\operatorname{TopK}(S, f, k)$ that returns the $k$ elements of a set $S$ with the largest values under a scoring function $f$. We define the \textsc{Core} set as the top fraction of communities by influence,
\begin{equation}
\label{eq:core-set}
\mathcal{C}_{\mathrm{core}}(q)=\operatorname{TopK}\big(\mathcal{C}(q),\ I,\ k_{\mathrm{core}}\big),
\end{equation}
where $k_{\mathrm{core}}=\max\{5,\,0.3\,|\mathcal{C}(q)|\}$. Similarly, we define the \textsc{Bridge} set as the top fraction by betweenness,
\begin{equation}
\label{eq:bridge-set}
\mathcal{C}_{\mathrm{bridge}}(q)=\operatorname{TopK}\big(\mathcal{C}(q),\ \operatorname{BC},\ k_{\mathrm{bridge}}\big),
\end{equation}
where $k_{\mathrm{bridge}}=\max\{3,\,0.2\,|\mathcal{C}(q)|\}$.
To identify \textsc{Satellite} themes, we include any remaining community that is strongly linked to a \textsc{Core} or \textsc{Bridge} theme in $H(q)$.
A community $c$ is marked \textsc{Satellite} if it has a strong meta-edge to any \textsc{Core} or \textsc{Bridge} community, i.e., if there exists $c'\in \mathcal{C}_{\mathrm{core}}(q)\cup \mathcal{C}_{\mathrm{bridge}}(q)$ such that $\max\{w(c,c'),w(c',c)\}\ge \tau_q$, where $w(\cdot,\cdot)$ is the meta-edge weight and $\tau_q$ is set to the median meta-edge weight for query $q$.
These rules are lightweight, stable across queries, and make role assignments interpretable.
These roles bias downstream packet selection toward salient themes and structurally meaningful cross-theme links while preserving enough peripheral context for realistic questions.

\subsection{Benchmark Instance Construction}
\label{sec:instance-construction}

Given the query-conditioned thematic structure produced in \S\ref{sec:community-roles}, the remaining steps construct a benchmark instance by (i) selecting explicit cross-theme links, (ii) packaging required information for cross-document sensemaking, and (iii) generating and verifying standalone QA pairs.

\subsubsection{Connector Relations}
\label{sec:connectors}

Sensemaking questions depend on multiple themes and on \emph{how those themes connect}. We therefore extract explicit connector relations that tie communities together. Let $\mathrm{comm}:V\rightarrow \mathcal{C}(q)$ map each entity to its community. An edge $e=(s,r,o,\mathrm{ev})\in E$ is a connector if it crosses community boundaries, i.e., $\mathrm{conn}(e)=\mathbb{I}\big[\mathrm{comm}(s)\neq \mathrm{comm}(o)\big]$. Not all cross-theme assertions are equally informative, so we retain a small top-$K$ set of connector edges ranked by their support signal (edge weight / evidence mass), favoring links that are repeatedly grounded in the retrieved corpus while keeping packets compact.

\subsubsection{Packet Construction}
\label{sec:packets}

To generate questions at scale without exposing the full graph or document set to an LLM, we construct compact \emph{packets} that preserve only the information required for cross-theme reasoning.

Each packet is built around a small bundle of themes that exposes a meaningful cross-theme linkage while remaining token-efficient. We begin with a high-influence \textsc{Core} theme and pair it with one or more \textsc{Bridge} themes that connect it to other regions of the meta-graph $H(q)$. In most cases, a bundle contains two themes (a salient \textsc{Core} and a connecting \textsc{Bridge}). When necessary to preserve a connector chain, we allow three-theme bundles (e.g., \textsc{Core}--\textsc{Bridge}--\textsc{Core}). A bundle is retained only if at least one selected connector relation links its themes, ensuring that resulting QA candidates depend on cross-theme connections rather than isolated fact lookups.

Given a selected bundle, a packet $P$ contains (i) each theme’s community card (summary and top findings with evidence IDs) and (ii) the connector relations linking themes in the bundle. This representation is intentionally compact, yet preserves the ingredients required for cross-document sensemaking.

\subsubsection{QA Generation and Verification}
\label{sec:qagen}

Conditioned on a packet $P(B)$, an LLM generator produces one-sentence, user-like questions with short, objective answers grounded in the provided findings and connector relations. The prompt biases toward realistic sensemaking intents (e.g., trigger, consequence, or cross-theme connection) and explicitly discourages trivia-like entity lookups (refer Prompt P1 in \S \ref{appendix}). We run deterministic decoding (temperature $0$), record all cited artifact IDs for traceability, and require the generator to output structured bookkeeping fields: required communities, supporting findings (at least one from each required community), and supporting connector IDs. These fields are not shown to models under evaluation, but they enable downstream filtering and analysis to ensure questions require cross-theme integration rather than a single isolated fact.

We filter candidates using an LLM-as-a-Judge paradigm \cite{zheng2024judging,gilardi2023chatgpt}. While reliability concerns are documented for subjective judgments \cite{huang2023chatgpt,gu2024survey,zheng2024judging}, prior work also finds that LLM judgments can reach agreement levels comparable to crowd-workers and, in some settings, match or exceed human performance \cite{gilardi2023chatgpt,guan2023language}. Our checks are intentionally evidence-only and objective: the judge must verify (a) that the candidate is supported by the provided artifacts and (b) that it \emph{necessarily} depends on multiple themes and at least one connector. This resembles fact verification and natural language inference, where LLMs have demonstrated strong performance \cite{guan2023language}. 

In our experiments, we employ a panel of three judge LLMs; each judge independently evaluates a candidate using the verifier prompt and returns a structured JSON decision with binary flags and short, artifact-grounded reasons (refer Prompt P2 in \S \ref{appendix}). A candidate is retained only if it satisfies all hard criteria and passes necessity tests. We aggregate judge outputs by majority vote and additionally reject candidates if any judge flags missing evidence, non-unique answers, or trivia drift, prioritizing precision over recall. 
\section{Dataset Characteristics and Artifacts}
\label{sec:dataset_characteristics}

An instance in \textsc{iAgentBench} is more than a QA pair: it includes auditable artifacts that reveal \emph{why} answering the question requires cross-document synthesis.
The design is guided by four goals: (G1) realism (interest-driven ODQA rather than KB- or trivia-induced questions), (G2) cross-theme dependence (questions that cannot be answered from a single theme), (G3) information-seeking fidelity (questions reflect common user intents), and (G4) auditability (traceable evidence and intermediate structure).
We reflect these goals in the released instance artifacts described below.

\paragraph{(G1) Realistic ODQA topics.}
Seed queries are sourced from real-world attention signals (\S\ref{sec:seed-queries}) and paired with query-conditioned web corpora, enabling evaluation on event-centric information needs that resemble what users search for.

\paragraph{(G2) Cross-theme, connector-dependent QA}
Each retained question is explicitly tied to at least two themes (communities) and at least one connector relation.
We release the supporting findings/connectors and their IDs so evaluators can diagnose whether a model failed due to retrieval or synthesis.

\paragraph{(G3) Information-seeking intent patterns.}
Beyond requiring cross-theme evidence, questions are designed to resemble how users express information needs when trying to make sense of an evolving topic. During generation we bias toward a small set of recurring intent patterns and record the intended pattern per instance as \texttt{intent\_pattern}. The five patterns capture common ways users seek understanding: \emph{explainer} asks for a grounded ``how/why'' that hinges on a specific link; \emph{connection} asks how two developments are related; \emph{trigger} asks what action or decision led to a downstream response; \emph{consequence} asks what outcome followed from a stated link; and \emph{stake} asks about a concrete condition, constraint, or objective tying developments together. This supports stratified analysis across intent types and evaluation beyond aggregate accuracy.

\paragraph{(G4) Auditability and released artifacts.}
Table~\ref{tab:dataset_fields} summarizes the core fields released per instance, including intent metadata that makes the information need explicit.
We additionally release intermediate artifacts to support analysis of graph-based retrieval and agent failures.
Instances can be regenerated for new time windows using the same construction rules, and because the retrieved corpora are recorded, the benchmark supports contamination checks and controlled re-runs.

\begin{table}[t]
\setlength{\tabcolsep}{5pt}
\renewcommand{\arraystretch}{1.08}
\centering
\footnotesize
\begin{tabularx}{\linewidth}{>{\raggedright\arraybackslash}p{0.30\linewidth} >{\raggedright\arraybackslash}X}
\toprule
{\bfseries Field} & {\bfseries Description}\\
\midrule
\texttt{id} & Unique identifier for the instance.\\
\texttt{query} & Seed query (interest-driven topic string).\\
\texttt{time\_window} & Time interval used for topic sourcing and retrieval.\\
\texttt{question} & Final one-sentence question.\\
\texttt{answer} & Short objective answer (entity, date, or short phrase).\\
\texttt{intent\_pattern} & Label for the information-seeking intent: \texttt{explainer}, \texttt{connection}, \texttt{trigger},\newline \texttt{consequence}, \texttt{stake}.\\

\texttt{retrieved\_sources} & URLs/metadata for retrieved documents in $\mathcal{D}(q)$.\\
\texttt{text\_units} & IDs of chunks/sentences used as evidence units.\\
\texttt{story\_graph\_stats} & Graph statistics for $G(q)$ (\#nodes, \#edges).\\
\texttt{communities} & Community IDs with role labels (core/bridge/satellite) and influence scores.\\
\texttt{community\_cards} & Per-community summary and grounded findings (with IDs).\\
\texttt{connectors} & Connector relations (with IDs and evidence pointers).\\
\texttt{packet} & Selected bundle $B$ and the packet $P(B)$ used for generation.\\
\texttt{supporting\_findings} & Finding IDs supporting the QA ($\ge 1$ per required community).\\
\texttt{supporting\_connectors} & Connector IDs required for the QA ($\ge 1$).\\
\texttt{judge\_decisions} & Outputs from the 3-LLM judge panel, including per-criterion flags and rationales.\\
\bottomrule
\end{tabularx}
\caption{Core fields released per \textsc{iAgentBench} instance.}
\label{tab:dataset_fields}
\end{table}
\section{Experiments \& Results}
\label{sec:results}
In this section, we evaluate how evidence access and evidence integration affect end-to-end QA performance, and how i\textsc{AgentBench} compares to standard ODQA benchmarks.
We compare four widely adopted LLMs under three inference settings: \emph{Base} (no external tools), \emph{RAG} (first page of retrieved documents from SearxNG \cite{searxng}), and \emph{Reflexion} (agentic self-reflection over retrieved evidence \cite{shinn2023reflexion}).
We report accuracy on two standard ODQA benchmarks, SimpleQA \cite{weiMeasuringShortformFactuality2024} and HotpotQA \cite{yangHotpotQADatasetDiverse2018a}, alongside \textsc{iAgentBench}.
Because these test sets are large and end-to-end agent evaluation is expensive, we randomly sample 500 questions from each benchmark, following similar approach in prior work \cite{sun2023think}.
Accuracy is computed using SimpleQA evaluation procedure \cite{weiMeasuringShortformFactuality2024}.

\begin{table*}[t]
\centering
\small
\begin{tabular}{lccccc|ccccc|ccccc}
\toprule
\multirow{2}{*}{\textbf{Model}} 
& \multicolumn{5}{c|}{\textbf{SimpleQA}} 
& \multicolumn{5}{c|}{\textbf{HotpotQA}} 
& \multicolumn{5}{c}{\textbf{\textsc{iAgentBench}}} \\
\cmidrule(lr){2-6} \cmidrule(lr){7-11} \cmidrule(lr){12-16}
& Base & RAG & Refl. & $\Delta_{\mathrm{RAG}}$ & $\Delta_{\mathrm{Refl}}$
& Base & RAG & Refl. & $\Delta_{\mathrm{RAG}}$ & $\Delta_{\mathrm{Refl}}$
& Base & RAG & Refl. & $\Delta_{\mathrm{RAG}}$ & $\Delta_{\mathrm{Refl}}$ \\
\midrule
Claude Sonnet 4.5            & 0.240 & \textbf{0.740} & 0.716 & 0.500 & $-$0.024 & 0.574 & 0.700 & \textbf{0.790} & 0.126 & 0.090 & 0.584 & 0.648 & \textbf{0.682} & 0.064 & 0.034 \\
LLaMA 4 Maverick 17B         & 0.174 & 0.657 & \textbf{0.826} & 0.483 & 0.169 & 0.456 & 0.537 & \textbf{0.758} & 0.081 & 0.221 & 0.356 & 0.532 & \textbf{0.628} & 0.176 & 0.096 \\
Mistral Large 3 (675B)       & 0.196 & 0.730 & \textbf{0.820} & 0.534 & 0.090 & 0.500 & 0.654 & \textbf{0.720} & 0.154 & 0.066 & 0.486 & \textbf{0.638} & 0.564 & 0.152 & $-$0.074 \\
Gemma 3 27B                  & 0.092 & 0.700 & \textbf{0.764} & 0.608 & 0.064 & 0.420 & 0.548 & \textbf{0.694} & 0.128 & 0.146 & 0.432 & \textbf{0.592} & 0.570 & 0.160 & $-$0.022 \\
\bottomrule
\end{tabular}
\caption{Accuracy of Base, RAG, \& Reflexion across SimpleQA, HotpotQA, and \textsc{iAgentBench}; $\Delta_{\mathrm{RAG}}=\mathrm{RAG}-\mathrm{Base}$, $\Delta_{\mathrm{Refl}}=\mathrm{Refl}-\mathrm{RAG}$.}

\label{tab:main_results}
\end{table*}

\begin{figure}[t]
\centering
\includegraphics[width=0.8\columnwidth]{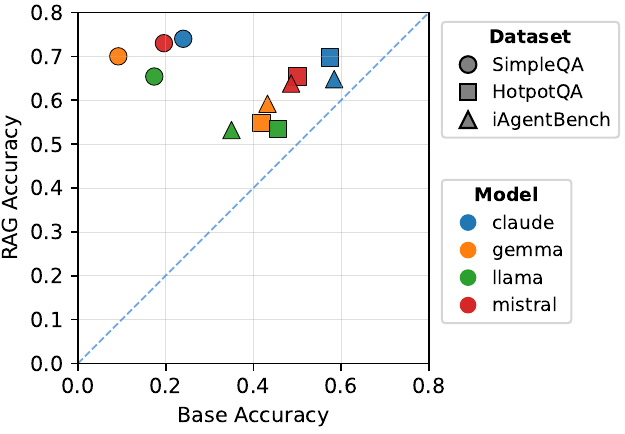}
\caption{Base vs. RAG accuracy across datasets and models. Points above the diagonal indicate gains from evidence access.}
\label{fig:base_vs_rag}
\end{figure}

\begin{figure}[t]
\centering
\includegraphics[width=0.8\columnwidth]{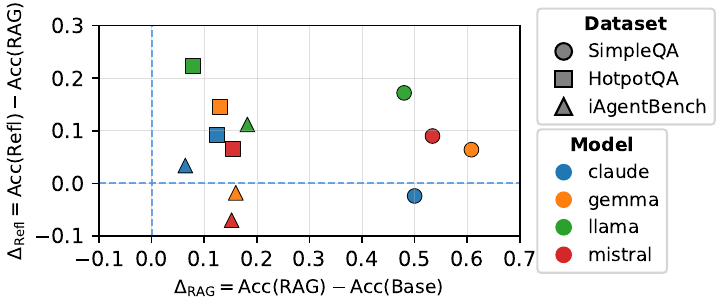}
\caption{Retrieval gains vs. agentic gains, decomposing improvements into $\Delta_{\mathrm{RAG}}=\mathrm{Acc}(\mathrm{RAG})-\mathrm{Acc}(\mathrm{Base})$ and $\Delta_{\mathrm{Refl}}=\mathrm{Acc}(\mathrm{Refl})-\mathrm{Acc}(\mathrm{RAG})$. Positive $\Delta_{\mathrm{Refl}}$ means iteration helps beyond RAG, while negative values indicate regressions.}
\label{fig:delta_rag_vs_reflexion}
\end{figure}

A first consistent pattern is that retrieval substantially improves performance across all datasets and models. In Figure~\ref{fig:base_vs_rag}, all points lie above the $y{=}x$ line, indicating that access to external evidence is beneficial even for strong frontier models. The magnitude of this effect differs across benchmarks. SimpleQA exhibits the largest Base$\rightarrow$RAG jumps, suggesting that many questions are primarily bottlenecked by evidence access: once a relevant passage is retrieved, models can often extract the short answer reliably. HotpotQA also benefits from retrieval, but the gains are more modest, consistent with a setting where evidence access and multi-hop composition jointly matter. \textsc{iAgentBench} likewise improves under RAG across the completed runs, but its remaining gap indicates that retrieving evidence is only part of the challenge and that cross-theme integration remains important.

These differences also reflect how the benchmarks are constructed and what they are designed to stress. SimpleQA is constructed to be challenging for base LLMs: it is adversarially collected against GPT-4-family responses and constrained to short questions with a single, indisputable answer~\cite{weiMeasuringShortformFactuality2024}. By design, this yields low Base accuracy across models, even though the questions are drawn from a broad set of topics. As seen in Table~\ref{tab:main_results}, once retrieval is enabled, SimpleQA becomes the easiest benchmark for most models, suggesting that its remaining difficulty is largely evidence access and precise extraction. In comparison, \textsc{iAgentBench} targets realistic, high-traffic topics drawn from attention signals and retains a sizable gap even under RAG, indicating that simply supplying evidence is not sufficient; systems must integrate information coherently across multiple pieces of context.

Figure~\ref{fig:delta_rag_vs_reflexion} further disentangles retrieval gains from the additional effect of multi-step self-reflection. While $\Delta_{\mathrm{RAG}}$ is typically positive, $\Delta_{\mathrm{Refl}}$ is mixed: Reflexion can improve over RAG when extra steps help connect dispersed evidence, but it can also regress when multi-step reasoning introduces drift or over-correction. This behavior is visible across datasets and is especially salient for \textsc{iAgentBench}, where models differ in whether iteration helps or hurts.

The table highlights these differences concretely. On SimpleQA, several models show large gains from Reflexion beyond RAG (e.g., LLaMA and Mistral), suggesting that iterative checking can help recover from initial extraction errors even when retrieval is strong. On HotpotQA, Reflexion yields a sizable boost for LLaMA, consistent with benefits for multi-hop composition, while improvements for other models are smaller. On \textsc{iAgentBench}, LLaMA benefits from Reflexion, but Mistral and Gemma decrease relative to RAG, reinforcing that agentic pipelines are not uniformly beneficial and should be evaluated for stability of evidence use rather than assuming monotonic improvements with more steps.

\section{Conclusion}
In this work, we present \textsc{iAgentBench}, a dynamic open-domain benchmark for evaluating information-seeking agents on questions that require cross-document sensemaking rather than single-passage extraction. \textsc{iAgentBench} couples traffic-driven topic selection with query-conditioned web corpora and a story-graph representation that exposes themes and the explicit links between them, enabling the construction of standalone, user-like questions that necessarily depend on multiple themes and connector relations. To support scientific use beyond headline accuracy, each instance is released with auditable artifacts, including intent patterns, supporting findings/connectors, packets, and judge decisions, so failures can be analyzed as breakdowns in evidence access, evidence integration, or multi-step stability.

Our experiments show that retrieval reliably improves performance across benchmarks, but \textsc{iAgentBench} retains a meaningful gap under RAG, indicating that evidence access alone does not resolve cross-theme questions. Decomposing improvements further reveals that agentic self-reflection is not uniformly beneficial; multi-step reasoning can help some models while degrading others, underscoring the need to evaluate not only retrieval quality but also the reliability of evidence use. We hope \textsc{iAgentBench} and its released artifacts provide a practical foundation for measuring and improving sensemaking-oriented information-seeking systems under evolving, real-world topics.

\section{Limitations}

\label{sec:limitations}

\paragraph{Cost of dynamic construction.}
Dynamic benchmarking increases cost relative to static QA datasets. \textsc{iAgentBench} requires retrieval, query-conditioned graph construction, and generation whenever a new evaluation window is produced, increasing compute and API overhead. This is an inherent trade-off of evaluating retrieval-augmented and agentic systems under evolving information.

\paragraph{Dependence on generative models.}
The pipeline relies on generative models for graph construction, question generation, and LLM-as-judge verification, so errors can arise from hallucinations, omissions, or biased evidence interpretation. We mitigate these risks with structured prompts, a multi-model judge panel, and by releasing intermediate artifacts for auditability. Prior work finds that for objective NLP assessment tasks, LLM-based annotation can reach agreement comparable to crowd-workers and, in some settings, match or exceed human performance \cite{gilardi2023chatgpt,guan2023language}.

\paragraph{Experimental coverage.}
We evaluate multiple LLM backbones under Base, RAG, and an agentic self-reflection variant, but alternative retrievers, tool sets, longer-horizon planning strategies, and broader hyperparameter sweeps remain to be explored. Expanding coverage is valuable but quickly becomes combinatorial and cost-prohibitive, so the current results are intended as a representative slice that enables comparable follow-on evaluations.


\bibliographystyle{ACM-Reference-Format}
\bibliography{ISAbench_references}

\clearpage
\onecolumn

\section{Appendix}
\label{appendix}

\subsection*{Prompts}

Here, we provide the prompts used in the \textsc{iAgentBench} framework for the QA generation step (P1) and the evaluation step (P2), with P2 used in the LLM-as-a-Judge setup.

\begin{figure*}[h]

\centering

\begin{adjustbox}{max width=\textwidth, max totalheight=0.75\textheight, keepaspectratio}
\begin{minipage}{\textwidth}
\begin{tcolorbox}[
    width=\textwidth,
    colback=lightgray!20, 
    colframe=gray, 
    title={\textbf{Prompt P2: LLM-as-a-Judge Verification Prompt}}, 
    fonttitle=\bfseries,
    sharp corners=southwest,
    coltitle=black,
    label=prompt:p2
]

\small 

SYSTEM: \newline
You are a EXPERT VERIFIER. Decide whether the ONE QA candidate should PASS or FAIL based ONLY on the evidence provided.

PASS criteria (all must hold):
\begin{enumerate}
    \item Evidence-only support:
    \begin{itemize}
        \item The answer is fully supported by the evidence texts provided inside the QA candidate object.
        \item No external knowledge, assumptions, or unstated facts may be used.
    \end{itemize}
    \item Multi-community necessity:
    \begin{itemize}
        \item The QA truly requires at least TWO communities.
        \item Necessity test: if you remove ALL finding texts from ANY ONE of the required communities, the question becomes unanswerable or clearly ambiguous.
    \end{itemize}
    \item Connector necessity:
    \begin{itemize}
        \item The QA truly requires at least ONE connector relation.
        \item Necessity test: if you remove ALL cited connector texts, the question becomes unanswerable or clearly ambiguous.
        \item The cited connector(s) must be necessary for deriving the answer (not just loosely related).
    \end{itemize}
    \item Objective QA:
    \begin{itemize}
        \item The question has a single best answer supported by the evidence.
        \item The question/answer are not subjective, interpretive, or perspective-dependent.
        \item The answer contains no hedging (``might/could/likely'') and no normative language (``good/bad/important'').
    \end{itemize}
    \item Natural user question:
    \begin{itemize}
        \item The question is exactly ONE sentence and reads like a natural user query (not a quiz/test item).
    \end{itemize}
    \item Anti-trivia:
    \begin{itemize}
        \item The QA is not answerable using only identity/alias/name equivalence.
        \item If identity/alias content is present, it must not be sufficient on its own; the QA must still depend on non-identity cross-community content.
    \end{itemize}
    \item Evidence presence and consistency:
    \begin{itemize}
        \item evidence\_findings contains $\geq$1 finding from EACH of at least TWO distinct community\_id values.
        \item evidence\_connectors contains $\geq$1 connector.
        \item Each evidence item must include both its ID(s) and its text.
    \end{itemize}
    \item Standalone clarity:
    \begin{itemize}
        \item The question must be fully interpretable and specific when read by itself, without needing any additional artifacts.
        \item All key referents must be explicit or uniquely identifying in the question.
        \item FAIL if a reasonable reader could ask ``which one?'' / ``who?'' / ``what exactly?'' due to missing antecedents or unspecified entities/events.
    \end{itemize}
\end{enumerate}

Inputs: \newline
QA\_CANDIDATE: \newline
\texttt{\{QA\_CANDIDATE\_JSON\}}

Output: \newline
Return ONE JSON object with EXACTLY these keys:
\begin{itemize}
    \item evidence\_only\_support\_flag: true/false
    \item evidence\_only\_support\_reasoning: string
    \item multi\_community\_necessity\_flag: true/false
    \item multi\_community\_necessity\_reasoning: string
    \item connector\_necessity\_flag: true/false
    \item connector\_necessity\_reasoning: string
    \item objective\_qa\_flag: true/false
    \item objective\_qa\_reasoning: string
    \item natural\_user\_question\_flag: true/false
    \item natural\_user\_question\_reasoning: string
    \item anti\_trivia\_flag: true/false
    \item anti\_trivia\_reasoning: string
    \item evidence\_presence\_consistency\_flag: true/false
    \item evidence\_presence\_consistency\_reasoning: string
    \item standalone\_clarity\_flag: true/false
    \item standalone\_clarity\_reasoning: string
\end{itemize}

Rules:
\begin{itemize}
    \item Each *\_reasoning must be 1--2 sentences, concrete, and refer only to QA\_CANDIDATE fields.
    \item If you cannot confidently mark a criterion as true based on QA\_CANDIDATE alone, set its *\_flag to false and explain why.
\end{itemize}

Return only the JSON object, no other text or markdown.

\end{tcolorbox}
\end{minipage}
\end{adjustbox}

\end{figure*}

\begin{figure*}[t]
\centering
\label{prompt:p1}

\begin{adjustbox}{max width=\textwidth, max totalheight=1\textheight, keepaspectratio}
\begin{minipage}{\textwidth}
\begin{tcolorbox}[
    width=\textwidth,
    colback=lightgray!20, 
    colframe=gray, 
    title={\textbf{Prompt P1: Q\&A Generation Prompt}}, 
    fonttitle=\bfseries,
    sharp corners=southwest,
    coltitle=black,
    label=prompt:p1
]

\small 

SYSTEM: \newline
You generate open-domain QA pairs that require CROSS-COMMUNITY SENSEMAKING over the provided STORY ARTIFACTS.

What you see (and ONLY what you can use):
\begin{itemize}
    \item COMMUNITY\_CARDS: community summaries + grounded findings (with IDs).
    \item CONNECTOR\_RELATIONS: explicit cross-community relations (with IDs) linking communities.
\end{itemize}

Definition: \newline
A cross-community sensemaking question cannot be answered using a single community card alone (or one isolated finding). \newline
It MUST require integrating information from MULTIPLE communities AND at least ONE explicit connector relation that links those communities.

Required constraints (must satisfy for every item):
\begin{enumerate}
    \item Multi-community requirement:
    \begin{itemize}
        \item Each QA MUST require at least TWO distinct communities.
        \item supporting\_findings MUST include $\geq$1 finding from EACH required community.
    \end{itemize}
    \item Connector requirement:
    \begin{itemize}
        \item Each QA MUST require $\geq$1 connector relation.
        \item The cited connector(s) MUST be necessary: if the cited connectors were removed, the question becomes unanswerable or ambiguous.
        \item The cited connector(s) MUST link the required communities (directly or via the stated relation).
    \end{itemize}
    \item Objective-answer requirement:
    \begin{itemize}
        \item The question MUST have a single best answer supported by the provided findings/connectors.
        \item Do NOT generate subjective, interpretive, or perspective-dependent questions or answers.
    \end{itemize}
\end{enumerate}

Question style (ONE sentence, realistic, user-like):
\begin{itemize}
    \item Sound like a real user trying to understand a situation (not a quiz).
    \item Prefer grounded sensemaking intents that hinge on explicit cross-community links:
    connection, trigger→response, cause→consequence, mechanism/role, stated condition/constraint.
    \item Good patterns include:
    \begin{itemize}
        \item ``How is X connected to Y?''
        \item ``What action/decision led to Y?''
        \item ``What triggered Y after X?''
        \item ``How did X contribute to Y?''
        \item ``What stated condition/demand tied X and Y together?''
    \end{itemize}
    \item Do NOT mention graphs, hypergraphs, communities, summaries, findings, packets, datasets, or ``according to the graph.''
    \item Avoid exam/trivia tone (e.g., ``Which leader...'', ``What country...'', ``In what year...'') unless it still clearly requires multi-community + connector dependence.
\end{itemize}

Do NOT generate:
\begin{itemize}
    \item Subjective/interpretive prompts: ``significance,'' ``importance,'' ``what it reveals,'' ``how it reflects,'' ``was it justified,'' ``who was right,'' ``how people felt,'' ``what it meant.''
    \item Anything whose answer could vary by perspective, framing, or values.
    \item Pure lookup / identity-only / alias-only questions (e.g., ``Who is X?'', ``What is X’s real name?''), unless the question ALSO requires a non-identity connector and genuine cross-community integration.
    \item Broad overview questions (``What happened?'', ``Summarize...'') that do not force a specific connector-dependent cross-community link.
\end{itemize}

Answer style:
\begin{itemize}
    \item Short, factual, uniquely supported by the artifacts (aim $\leq$ 12 tokens; slightly longer only if unavoidable).
    \item No hedging (``might/could/likely'') and no normative language (``good/bad/important'').
    \item Do NOT introduce facts not present in the provided findings/connectors.
    \item If a unique objective answer is not supported, do not output that QA pair.
\end{itemize}

Intent labels: \newline
Set intent\_pattern to one of \texttt{["explainer","connection","trigger","consequence","stake"]} where:
\begin{itemize}
    \item explainer: a grounded explanation that hinges on a specific cross-community link (not ``significance'')
    \item connection: how two developments/entities are related via connector(s)
    \item trigger: what action/decision/event led to a cross-community response/outcome
    \item consequence: what outcome followed from a cross-community link
    \item stake: a concrete stated constraint/condition/demand/objective tying communities (NOT ``why was it important'')
\end{itemize}

Internal self-checks (run these before outputting each item; do not output the check text):
\begin{enumerate}
    \item Objective-answer test:
    \begin{itemize}
        \item Is there exactly one best short answer supported by the cited findings/connectors?
        \item Would reasonable readers give the same answer?
    \end{itemize}
    \item Multi-community necessity test:
    \begin{itemize}
        \item If ANY one required community card is removed, does the question become unanswerable or ambiguous?
    \end{itemize}
    \item Connector necessity test:
    \begin{itemize}
        \item If ALL cited connector relations are removed, does the question become unanswerable or ambiguous?
    \end{itemize}
    \item Anti-trivia test:
    \begin{itemize}
        \item Does it feel like a real sensemaking question (not a quiz or entity-attribute lookup)?
    \end{itemize}
\end{enumerate}

Output: \newline
Return a JSON array of up to \texttt{\{K\}} items. Each item must be:
\begin{itemize}
    \item question: string (ONE sentence)
    \item answer: string (short, objective, supported)
    \item required\_communities: [2+ community\_id]
    \item supporting\_findings: [{community\_id: ..., finding\_id: ...}, ...]  ($\geq$1 per required community)
    \item supporting\_connectors: [connector\_id, ...] ($\geq$1; must be necessary; must link required communities)
    \item intent\_pattern: one of ["explainer","connection","trigger","consequence","stake"]
    \item why\_multi\_community: ONE sentence explaining why multiple communities + connector(s) are required
    \item (describe it in content terms; do NOT mention graphs/communities/packets)
\end{itemize}

Diversity:
\begin{itemize}
    \item Across the set, vary intent\_pattern when possible and avoid using the same connector\_id for all items unless unavoidable.
\end{itemize}

COMMUNITY\_CARDS: \newline
\texttt{\{COMMUNITY\_CARDS\_JSON\}}

CONNECTOR\_RELATIONS: \newline
\texttt{\{CONNECTORS\_WITH\_IDS\}}

Generate up to \texttt{\{K\}} QA pairs. Return ONLY the JSON array, with no extra text and no markdown.

\end{tcolorbox}
\end{minipage}
\end{adjustbox}

\end{figure*}

\end{document}